\documentclass[conference]{IEEEtran}
\ifCLASSINFOpdf
\else
\fi
\usepackage{graphicx} 
\usepackage{subcaption}
\usepackage[margin=1in]{geometry}
\usepackage[dvipsnames]{xcolor}
\usepackage{inputenc}
\usepackage{amsmath}
\usepackage{xcolor}
\usepackage{booktabs}
\usepackage{comment}
\usepackage{algorithm}
\usepackage{algpseudocode}

\begin{document}

\title{Meme Similarity and Emotion Detection using Multimodal Analysis}


\author{\IEEEauthorblockN{Aidos Konyspay,
Pakizar Shamoi \IEEEauthorrefmark{1}, Malika Ziyada, Zhusup Smambayev}
\IEEEauthorblockA{School of Information Technology and Engineering \\
Kazakh-British Technical University\\
Almaty, Kazakhstan\\
Email: 
\IEEEauthorrefmark{1}p.shamoi@kbtu.kz
}
}

\maketitle

\IEEEpeerreviewmaketitle
\begin{abstract}
Internet memes are a central element of online culture, blending images and text. While substantial research has focused on either the visual or textual components of memes, little attention has been given to their interplay. This gap raises a key question: What methodology can effectively compare memes and the emotions they elicit? Our study employs a multimodal methodological approach, analyzing both the visual and textual elements of memes. Specifically, we perform a multimodal CLIP (Contrastive Language-Image Pre-training)  model for grouping similar memes based on text and visual content embeddings, enabling robust similarity assessments across modalities. Using the Reddit Meme Dataset and Memotion Dataset, we extract low-level visual features and high-level semantic features to identify similar meme pairs. To validate these automated similarity assessments, we conducted a user study with 50 participants, asking them to provide yes/no responses regarding meme similarity and their emotional reactions. The comparison of experimental results with human judgments showed a 67.23\% agreement, suggesting that the computational approach aligns well with human perception. Additionally, we implemented a text-based classifier using the DistilBERT model to categorize memes into one of six basic emotions. The results indicate that anger and joy are the dominant emotions in memes, with motivational memes eliciting stronger emotional responses.  This research contributes to the study of multimodal memes, enhancing both language-based and visual approaches to analyzing and improving online visual communication and user experiences. Furthermore, it provides insights for better content moderation strategies in online platforms.

\end{abstract}

\section{Introduction}
In today's digital age, Internet memes are one of the most popular ways to share emotions, ideas, and thoughts. For example, people shared more than one million memes daily on the Instagram platform in 2020 \cite{meme}. The authors of memes use text, images, and videos in creative ways to share ideas or jokes. Thus, memes serve as entertainment and stress relief sources while providing social meaning and highlighting important issues.


Memes have a unique ability to cross cultural boundaries, creating a shared language that many people understand and enjoy \cite{olena2020memes, Grishaeva2021}. As memes spread rapidly across social media, understanding their content and impact has become increasingly important. The meaning of memes can differ between countries and cultures. Analyzing these differences can help improve recommendation algorithms on social media platforms.


However, current sentiment analysis tools predominantly focus on textual or visual data independently, often overlooking the complex interplay between text and image that defines memes. This oversight hinders computers' ability to interpret memes in a manner similar to humans, thereby reducing the effectiveness of analyses in recommendation systems.



This research develops a model to group similar memes based on their visual and textual content and the emotions they evoke. By using CLIP and DistilBERT models, we aim to answer the following questions: 
\begin{itemize}
  \item How are the visual and textual features of memes related?
How can memes be effectively categorized into groups?
     
\item What emotions are commonly associated with particular groups of memes?
    
\end{itemize}

The contributions of this paper are as follows:
\begin{itemize}
    \item \textit{Simultaneous Grouping by Content.} The study develops a method to group memes by analyzing both their visual (what you see) and textual (what you read) content at the same time. This helps to better group memes based on their overall message.
    

    \item \textit{Understanding Meme Emotions.} The research explores the emotions memes evoke. By identifying emotions, we can better understand the feelings that memes induce. Memes are funny. However, they contain humor in various forms of emotions. For instance, motivational memes could provoke feelings of joy, anger, or even sadness.  This emotional diversity in memes remains underexplored in the existing literature. 
    
    
    \item \textit{Expanding the Memotion Dataset. } Our study extends the existing Memotion dataset by incorporating the results of our six-class emotion analysis. Specifically, we classify memes into one of six emotions: sadness, joy, love, anger, fear, or surprise. This new addition is useful in enhancing the dataset with basic sentiments (positive, positive, negative, negative, neutral), humor, sarcasm, offense, and motivation.  For example, while the dataset pointed out that something is ‘motivational’, our updated annotations indicate the emotion used; it may be joy, anger, etc. This expansion makes the dataset richer and more useful for future studies. 
\end{itemize}





The paper is structured as follows. Section I is the Introduction. Section II reviews the Related Work. Section III details the methods used in this study, including data, grouping similar memes, emotion detection in memes, CLIP, and cosine similarity. The results are presented in Section IV. Section V discusses the limitations and future work, followed by the conclusion in Section VI.

\section{Related Work}

In our digital age, memes have become a powerful way to share laughs, opinions, and even political stances. But no matter how carefree they may seem, some memes have a negative side. For example, they could contain offensive or harmful content. Researchers use a variety of methods to analyze the appeal of memes and identify those that are not ethically appropriate. 

\subsection{
Image-based Analysis of Memes
}
The absence of textual headers in memes makes image-based grouping a challenging task for researchers. For example, a recent study focused on image-based meme classification using convolutional and transformer architectures. The study showed that visual data alone can achieve high accuracy, but meme analysis often depends on cultural context \cite{Gao2024}.

Meme studies mainly focus on images because most memes include visual attributes. A recent study used Deep Convolutional Neural Networks (CNNs) to explore object and visual features in memes. It was the first to show their role in creating memes and recognizing humor \cite{Choi2023}. Similarly, some researchers used YOLOv5 to detect objects in meme images, improving meme categorization based on content \cite{Wang2024}. This is a new direction, which introduced the use of GANs for generating memes and classifying meme emotions \cite{Sharma2022}. This paper reviewed studies on image detection techniques that enhance meme categorization and sentiment analysis.

\subsection{
Text-based Analysis of Memes
}

Recent meme analysis focused on text-based approaches, as key features helped reveal the meme's meaning and cultural context. One study explored language features in contextual memes and used deep learning to examine how slang and humor influence meme valency. This model improved meme text analysis by 33\% using a structured linguistic model \cite{Xu2022}.

Similarly, another study introduced a transformer for sentiment analysis of meme texts, incorporating irony and sarcasm. 
They applied their approach to improving sentiment classification in memes, outperforming traditional models \cite{Kim2023}. The study \cite{Bennett2021} used the SS method to detect offensive language in meme captions, contributing to automatic hate speech detection.

Furthermore,   the study \cite{Pang2023} extended the analysis of cultural references to meme texts. They discovered that text-based meme analysis uncovers the hidden cultural semiotics of memes. This approach examines memes from a cultural perspective using only textual features.  Next, \cite{Rajendran2022} presented the latest meme text summarization system. Their method extracted memes' semantic meaning using natural language processing (NLP) tools, focusing on textual content without considering contextual pictures.

New trends in meme analysis have been developed in connection with the creation of large datasets designed for the training and evaluation of AI systems. The study \cite{Guillermo2020Infotec} presented an analysis of deep learning and text categorization for meme classification via the \textit{SemEval-2020} shared activity.

\subsection{
Multimodal Analysis of Memes
}
A recent trend in meme study is multimodal analysis, which examines both the verbal and visual elements of memes. This enables a richer analysis of memes and the variety of emotions that they can convey. Recent techniques such as CLIP performed well in understanding the combination of text and photographs within memes \cite{Zhang2024Long, Kumar2022Hate}. These methods extended contrastive learning to classify memes using both visual and textual features, increasing the number of correctly classified memes.

Several studies explored how self-supervision and multimodal analysis can handle large volumes of unlabeled memes. For instance, \cite{Tung2023} introduced \textit{SemiMemes}, a semi-supervised model that improved meme classification using deep learning on unlabeled data. This approach adopts a bimodal approach, considering both the image and text as distinct yet related pieces of information. \cite{Alluri2021} analyzed multimodal sentiment-capturing transformers to categorize memes as ironic, humorous, or motivational. Subsequently, advanced large language models (LLMs) improved unsupervised multimodal topic modeling for meme classification \cite{Prakash2023}. Next, \cite{Cui2023} contributed a new contrastive learning approach to meme category detection. This approach used "Be-or-Not" prompts to generate hard negative samples, enhancing the ability to distinguish subtle meme categories.


In \cite{Lin2023}, the author tackles the challenge of using LLMs for multimodal reasoning in the detection of dangerous memes. This work proposes new methods to improve harm detection in memes and related datasets. Additionally, \cite{Koutlis2023} introduced \textit{Meme-Tector}, which improves meme detection by combining visual elements and attention mechanisms for better reliability.

Numerous papers examined the topic of meme retrieval and search systems. For example, the \cite{simmeme} proposed a \textit{SimMeme}, a graph-based search engine that uses a graph similarity measure for enhanced meme analysis. Unlike earlier image search methods, it accounts for image similarity, user tags, and meme relationships.

\subsection{
Meme Sentiment, Emotion, and Toxicity Detection
}

There is an increasing interest in identifying potentially toxic elements of memes because humor, irony, and sarcasm often carry offensive meanings. Extensive research has been conducted using deep learning and multimodal reasoning to address this topic.  \cite{Ma2022} have formulated a multi-task learning algorithm combining textual and visual classification to better detect hateful memes. Similarly, \cite{Pimpalkar2022} highlighted the importance of using machine learning to recognize sentiment in visual-textual memes and analyze hidden meanings.

Focus on contextualization of the meme within social media is also crucial. For example, \cite{Joshi2023} used knowledge graphs to track and analyze memes in both Reddit and Discord. \cite{Joshi2023} analyzed the sentiment of memes and their allocation between subreddits or Discord groups.

The problem of toxic memes was addressed in \cite{Qu2022}, where the authors proposed a multimodal contrastive learning approach to analyze hate speech memes. Their work explained how memes can transform and become a source of provocation and sexual harassment in a certain period. Offensive meme detection was explored in works like \cite{Giri2021}, which employed NLP techniques to classify offensive content and assess toxicity levels.

Next, \cite{Gupta2021} applied multimodal sentiment analysis for memes, using self-supervised image in-painting to improve understanding of missing pixels. This work aims to address the growing complexity of interpreting visual-textual relations in memes, particularly negative ones.

Thus, several studies employed an opinion-mining approach to analyze the meme. While \cite{hatzivassiloglou1997predicting}, \cite{hu2004mining}, and \cite{taboada2006methods} analyzed only adjectives for sentiment polarity calculation, the authors in \cite{Amalia2018Meme} considered all parts of speech for the meme sentiment analysis. They applied Optical Character Recognition (OCR) to extract captions from memes and used the Naïve Bayes algorithm to classify the memes as positive or negative.

Recent studies have also expanded multimodal learning by integrating methods like CLIP to enhance image-text classification. For instance, \cite{Burbi2023} used a CLIP vision-language model to improve hateful meme classification by mapping meme semantics to words. Similarly, \cite{Jannat2022} focused on sentiment classification in multimodal Bengali memes, highlighting the challenges of sentiment analysis across languages. 

The analysis of sentiments in memes depends on how people perceive emotions through visual elements, as reported in \cite{10652090}. Detecting sentiment in memes requires advanced systems because behavioral research demonstrates that multiple cues help users recognize emotions, according to \cite{Sakai2020}. The sentiment analysis of social contexts benefits from fine-tuned models according to research in stress detection and behavioral computing studies \cite{RAMACHANDRA202411, Tazin2024, Michele2024, 10412045}.

Fine-grained meme understanding seeks to interpret memes across multiple tasks, such as sentiment and offensiveness detection, by capturing interactions between text and image modalities. The proposed Metaphor-aware Multi-modal Multi-task Framework (MF) addresses existing limitations by modeling inter- and intra-modality interactions, as well as task correlations, demonstrating better performance on the MET-Meme dataset compared to current baselines\cite{WANG2024111778}. 

The increasing use of memes on social media to express opinions and measure public sentiment highlights the need for models that capture their affective dimensions. Current approaches struggle with generalization due to limited visual-linguistic grounding. Recently proposed ALFRED framework, utilizing the MOOD dataset, demonstrates improved emotion detection in memes through effective cross-modal fusion and emotion-aware visual cue modeling \cite{meme_emotion_new}.  Additionally, \cite{memechat} presented a model that integrates memes into general open-domain dialogue systems, specifically for meme selection. As shown in \cite{memechat}, integrating memes into dialogue systems enhances text matching, supporting meme search, and better user engagement.

\subsection{
Cultural and Sociological Perspectives on Memes
}

Memes reflect cultural elements, including morals, politics, and debates. Many researchers have investigated memes' role as a medium for intercultural communication. 

\cite{Weng2014Predicting} developed the model to predict meme popularity based on the communities, showing how memes are disseminated and transformed across virtual groups. Moreover, this study examines the dynamics between traditional news topics and social media content, highlighting how topical relevance influences social media popularity. By analyzing New York Times topics alongside image-text memes from Reddit, it finds that topic alignment with news trends boosts meme virality, with early-topic memes showing notable predictive power for popularity using topicality features in a CatBoost classifier \cite{Barnes2024}.

\cite{Ferrara2013Clustering} and \cite{Jesus2020Semantic} explored how content and metadata can enable memetic clustering to find memes by meaning. These approaches help explain the shared subjectivity of memes and their potential to shape culture. \cite{zannettou2018origins} studied memes as cultural phenomena, examining how subcultures create memes to influence broader culture, showing both the positive and negative aspects of memes' role in shaping collective identity.

It is common for political memes to include a great deal of text content; however, most of them are very visually powerful.  \cite{Singh2023} studied how political memes could be identified using image-only approaches. They concluded that symbols and color can help classify political meme intent.

\subsection{
Current Challenges and Limitations
}


As can be seen, numerous theoretical and empirical advancements have been made since meme analysis first began. Multimodal approaches and self-supervised learning were helpful in meme classification and sentiment analysis. By using various approaches, researchers are increasingly highlighting memes as both humorous and problematic objects. Subsequent studies on feature extraction and multimodal analysis are crucial for advancing meme classification. \cite{Smitha2018Meme} used both text and image features, while \cite{Cochrane2022} introduced a two-part classification system for VT memes. These methods, along with some others, such as CLIP, have improved the ability to classify all of the different types of memes.

Nevertheless, several important limitations are evident in meme analysis. Image processing methods like CNNs and YOLOv5 are sensitive to datasets, reducing efficiency when applied to different meme styles or cultures. Moreover, text-based approaches are ineffective for meme text due to its informal and context-sensitive nature. Even modern systems that detect sarcasm and sentiment face challenges with slang, regional languages, and cultural references, making it difficult to identify meme content. These models often fail to analyze the crucial interaction between text and image, which is essential for understanding a meme's full tone and meaning. There is a need for future research that develops more efficient resource models that are context-sensitive and can capture the richness of meme analysis.

\section{Methods}
\subsection{Proposed Approach Overview}
\begin{figure}[tb]
    \centering
    \includegraphics[width=0.5\textwidth]{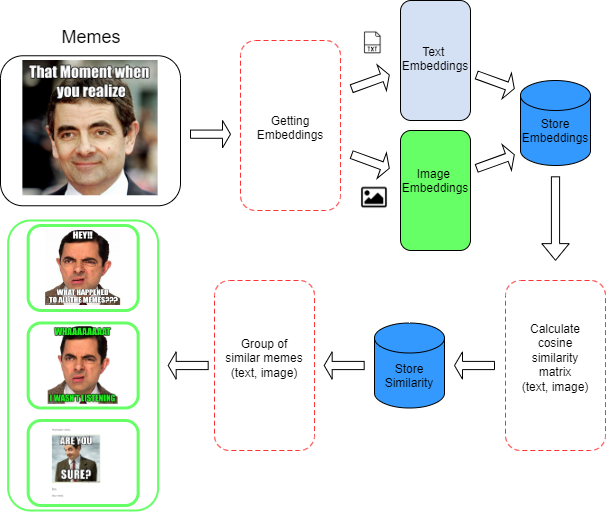}
    \caption{Proposed Approach for Grouping Similar Internet Memes
Using CLIP}
\label{main}
\end{figure}
In this paper, we aim to group memes and identify the emotions that they evoke. The general approach is illustrated in Fig \ref{main}. It includes several steps. Firstly, we need to get the embeddings for memes using the CLIP model. Secondly, we will calculate cosine similarity and compare CLIP embeddings to similar memes in groups. Finally, to retrieve meme emotion, we use DistilBERT.

Memes are a unique combination of texts and images, so consequently, their analysis requires a multimodal approach. To that end, we use the CLIP model to get meme embeddings. Such embeddings are beneficial as they encode different modes of memes as text and images in a single latent space.

\subsection{Data}
\begin{figure*}[tb]
    \centering
    \begin{subfigure}[b]{0.32\textwidth}
        \centering
        \includegraphics[width=\textwidth]{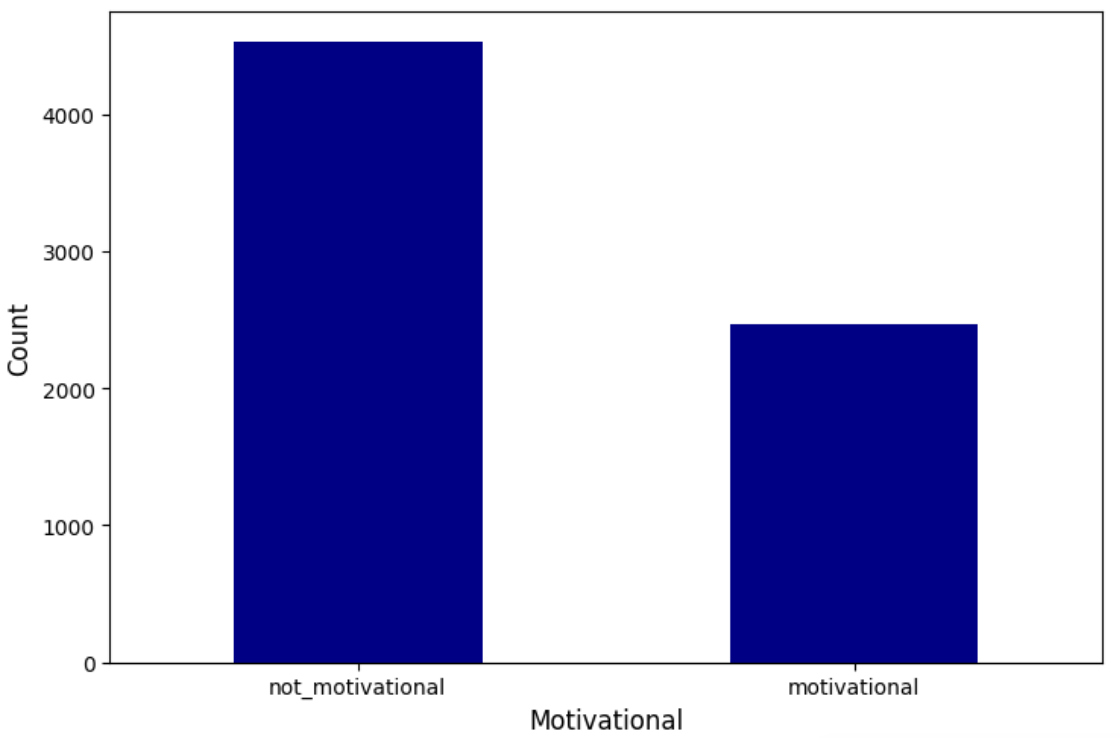}
        \caption{Distribution of Motivational Memes}
        \label{fig:humor}
    \end{subfigure}
    \hfill
    \begin{subfigure}[b]{0.32\textwidth}
        \centering
        \includegraphics[width=\textwidth]{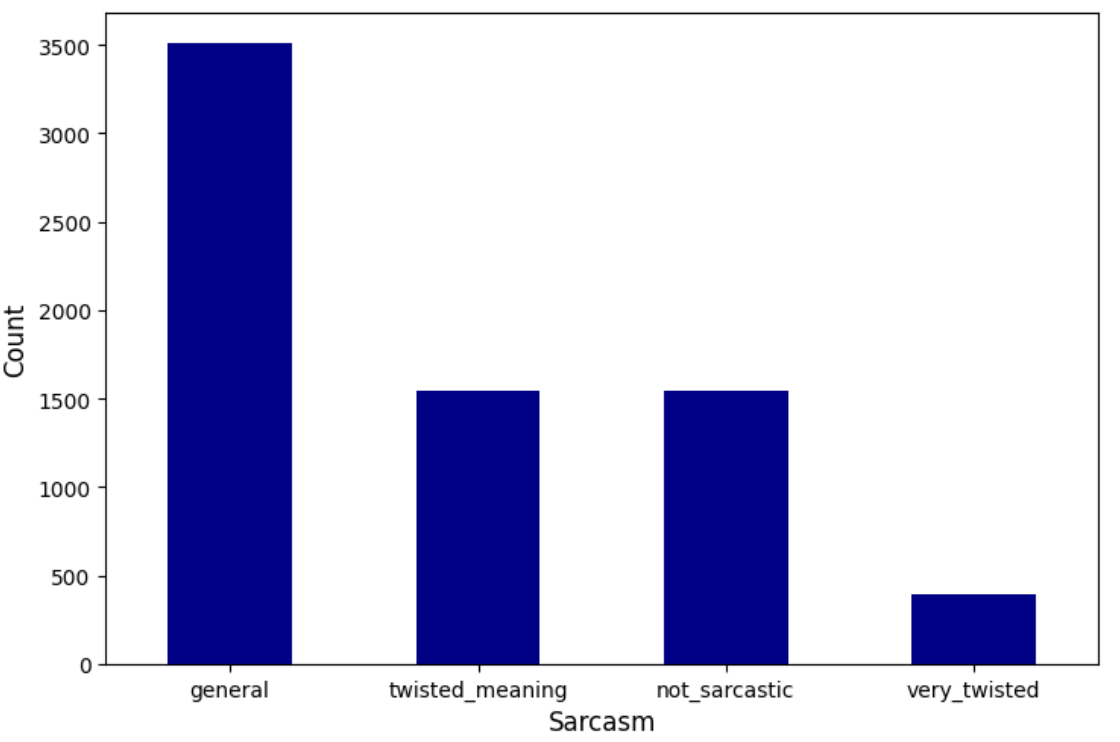}
        \caption{Distribution of Sarcasm}
        \label{fig:sarcasm}
    \end{subfigure}
    \hfill
    \begin{subfigure}[b]{0.32\textwidth}
        \centering
        \includegraphics[width=\textwidth]{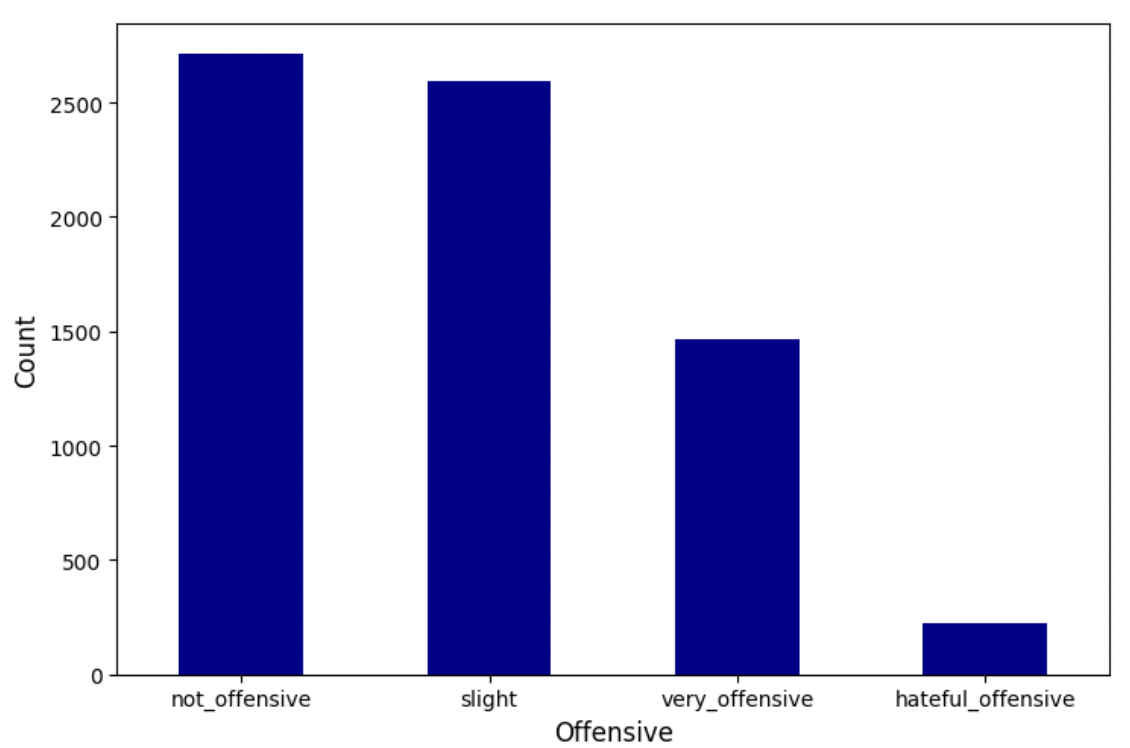}
        \caption{Distribution of Offensive Content}
        \label{fig:offensive}
    \end{subfigure}

    \vfill
    \begin{subfigure}[b]{0.48\textwidth}
        \centering
        \includegraphics[width=0.7\textwidth]{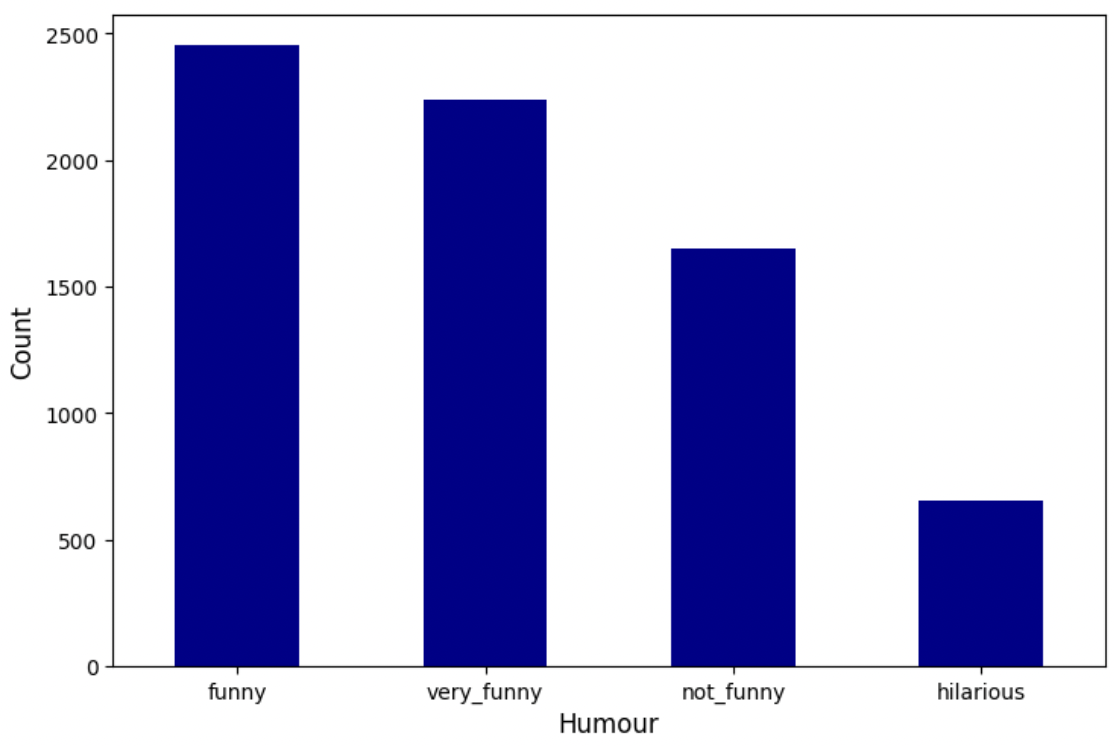}
        \caption{Distribution of Humour Content}
        \label{fig:motivational}
    \end{subfigure}
    \hfill
    \begin{subfigure}[b]{0.48\textwidth}
        \centering
        \includegraphics[width=0.7\textwidth]{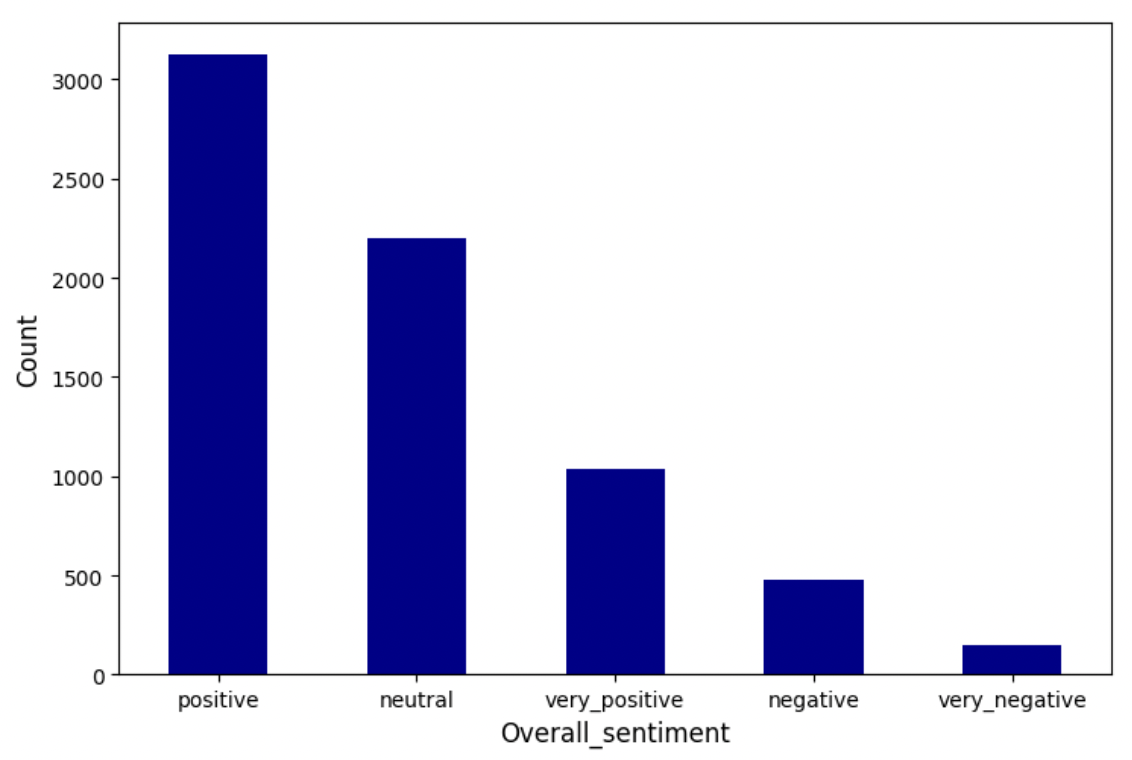}
        \caption{Distribution of Overall Sentiment}
        \label{fig:overall_sentiment}
    \end{subfigure}
    \hfill

    \caption{Bar charts illustrating the distribution of various attributes in the Memotion dataset.}
    \label{fig:distributions}
\end{figure*}

\begin{figure}[tb]
    \centering
    \includegraphics[width=0.5\textwidth]{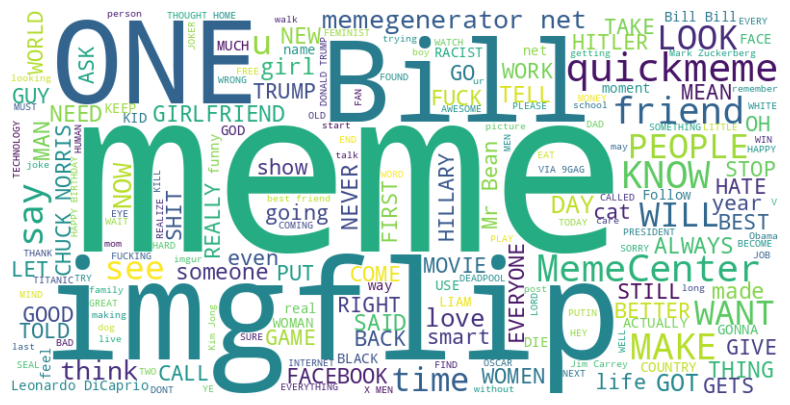}
    \caption{Word cloud of the text in Memotion dataset.}
    \label{wordcloud}
\end{figure}

For this research, we used a Memotion dataset \cite{chhavi2020memotion}, containing a total of 6,992 images. 
Each image is associated with humor, sarcasm, motivation, and overall sentiment values. Humor in each image is categorized as either 'funny' or 'not funny.' Similarly, sarcasm is labeled as 'sarcastic' or 'non-sarcastic', and motivation is marked as 'motivational' or 'non-motivational'. As for the overall sentiment, it's classified into three categories: 'positive', 'negative', and 'neutral'. The distribution of memes across these categories is shown in Fig. \ref{fig:distributions}. The word cloud of the texts in memes is presented in Fig. \ref{wordcloud}.

This dataset also contains columns such as the image name and the corrected text of the meme, which we will use in our experiment to get image and text embeddings.

We conducted an experiment to group similar memes by image and text using CLIP on the Memotion dataset. To ensure our results' validity and real-world relevance, we repeated the experiments on the Reddit Memes Dataset \cite{kaggleRedditMemes}, which includes 3226 images along with meme titles, images, and publication times.


\subsection{Grouping Similar Memes}

To group memes based on their similarity, we used the CLIP embeddings and performed pairwise comparisons between individual meme embeddings according to their cosine similarities (see Fig. \ref{main}). For two memes to be classified as similar, the cosine distance between their embeddings was used, and a threshold was used at 0.8. Memes with cosine similarity above this threshold were considered original, and those with higher similarity were grouped together. 
This threshold was chosen to achieve a balance between grouping closely related memes and preventing overly broad categorizations. The steps for grouping memes were as follows:
\begin{itemize}
    \item To get embeddings for each meme, use the CLIP model.
    The following step is to compute the cosine similarity of two embeddings of all meme pairs.
    \item Use a tolerance level of 0.8 when looking for similar meme pairs.
    All pairs are grouped iteratively to create clusters of similar, loudly related memes from the initial list of memes.
\end{itemize}

As a result of this process, we identified meme groups with similar textual and visual content.

\subsection{Contrastive Language-Image Pre-training}
We used CLIP \cite{Radford2021LearningTV} because it has revolutionized multi-modal learning by bridging the gap between text and images. This approach is advantageous for complex tasks like meme analysis, where both visual and textual understanding are key. The model combines image and text embeddings, trained on 400 million image-text pairs with self-supervised learning, mapping them to a shared space. For example, a cat image and the text "a picture of a cat" have similar coordinates in this space.

CLIP demonstrated impressive zero-shot capabilities and transferability across various tasks \cite{Zhou2023, Luo2022CLIP4Clip}. While initially developed for image-text pairs, CLIP has been adapted for video retrieval \cite{Luo2022CLIP4Clip}, region-based object detection \cite{Zhong2022RegionCLIP}, and even 3D point cloud understanding \cite{Zhang2022PointCLIP}. The model has shown promise in zero-shot vision-and-language navigation \cite{Dorbala2022CLIP} and enhanced image retrieval \cite{Lahajal2024Enhancing}. Recent advancements include Long-CLIP, which extends CLIP's capability to handle longer text inputs \cite{Zhang2024Long}, and CLIP4Clip, which transfers CLIP's image-text knowledge to video-text retrieval  \cite{Luo2022CLIP4Clip}. These developments highlight CLIP's potential for further applications in multimodal understanding tasks.

\begin{figure}[tb]
    \centering
    \includegraphics[width=0.5\textwidth]{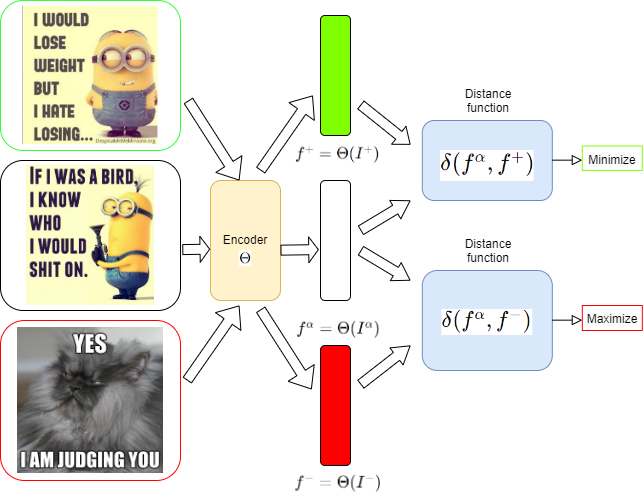}
    \caption{A visualization of the CLIP model minimizes the distance function between similar memes and maximizes it between different ones}
\label{clip}
\end{figure}

As you can see in Fig. \ref{clip}, CLIP trains both an image encoder (like a CNN, e.g., ResNet) and a text encoder (transformer model) together to match images with their correct text descriptions. Both the image and text encoders project their inputs into a shared embedding space. The aim is to bring related images and texts close together in this space.
CLIP aims to use contrastive loss to minimize the cosine distance for matching image-text embeddings and maximize it for non-matching pairs.

Given a batch of \( N \) image-text pairs \( \{(I_i, T_i)\}_{i=1}^{N} \), the contrastive loss for each pair \( (I_i, T_i) \) is computed as:
\begin{equation}
\mathcal{L}_i = -\log \frac{\exp(\text{sim}(I_i, T_i) / \tau)}{\sum_{j=1}^{N} \exp(\text{sim}(I_i, T_j) / \tau)}
\end{equation}
where \( \tau \) is a temperature parameter.
The total loss for the batch is the sum of the individual losses:
\begin{equation}
\mathcal{L} = \sum_{i=1}^{N} \mathcal{L}_i
\end{equation}

When used as a classifier on a new dataset, the pre-trained text encoder automatically sets up a linear classifier. It does this by creating embeddings of the new classes from their text descriptions.

\subsection{Cosine Similarity}
Cosine similarity is a useful tool to measure how similar the image and text features are to each other when using their numerical embeddings in a high-dimensional space.
We calculated the Cosine similarity coefficient using Eq. 3. Our model uses it to assess the degree of similarity between different meme embeddings. 
\begin{equation}
\small
\text{cosine similarity} = \frac{A \cdot B}{\|A\| \|B\|} = \frac{\sum_{i=1}^{n} A_i B_i}{\sqrt{\sum_{i=1}^{n} A_i^2} \sqrt{\sum_{i=1}^{n} B_i^2}}
\end{equation}

\begin{figure}[tb]
    \centering
    \includegraphics[width=0.5\textwidth]{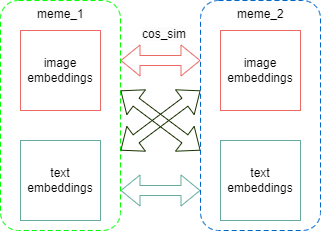}
    \caption{Visualization of calculation of the cosine distance between two memes}
    \label{vis}
\end{figure}

\begin{itemize}
    \item If two vectors are oriented in the same direction, their cosine similarity is 1 with an angle of 0 degrees between them.
    \item If the vectors are perpendicular, their cosine similarity is 0.
    \item When two vectors point in opposite directions (180-degree angle), their cosine similarity is -1.
\end{itemize}

In summary, smaller angles between vectors mean higher cosine similarity, indicating they are more alike. This metric is particularly useful for understanding semantic similarity in document analysis and has been integrated into our comparison of embedding techniques.

As shown in Fig. \ref{vis}, we compared memes by analyzing both the images and the text. We compared the image of the first meme with the image of the second meme and the text of the first meme with the text of the second meme. Additionally, we calculated the cosine similarity between the image of the first meme and the text of the second meme and vice versa for the second meme.

\subsection{Emotion Detection in Memes}

\begin{figure}[tb]
    \centering
    \includegraphics[width=0.5\textwidth]{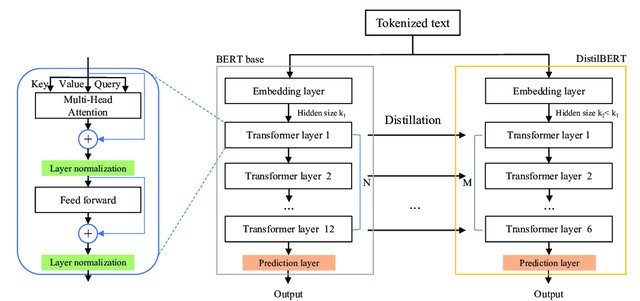}
    \caption{The DistilBERT model architecture \cite{Adel2022}}
\label{fig:bert_distilbert}
\end{figure}

Emotion detection in memes was performed using DistilBERT, a compact, fast, and efficient version of BERT. DistilBERT retains 97\% accuracy of BERT while significantly reducing the model size and output time \cite{Sanh2019DistilBERT}. It uses the same transformer architecture and learning objectives as BERT \cite{devlin2019bertpretrainingdeepbidirectional} but includes optimizations to improve efficiency (see Fig.~\ref{fig:bert_distilbert}).

\begin{figure}[htb]
    \centering
    {
    \begin{minipage}{0.5\textwidth}
        \centering
        \includegraphics[width=\textwidth]{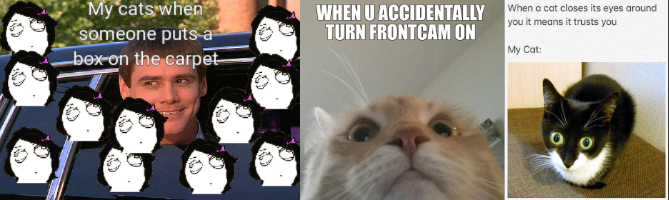}
    \end{minipage}

    \vspace{0.5cm} 
    
    \begin{minipage}{0.5\textwidth}
        \centering
        \includegraphics[width=\textwidth]{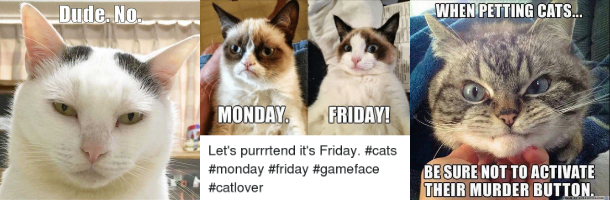}
    \end{minipage}
    }
    \vspace{0.5cm} 

    \begin{minipage}{0.5\textwidth}
        \centering
        \includegraphics[width=\textwidth]{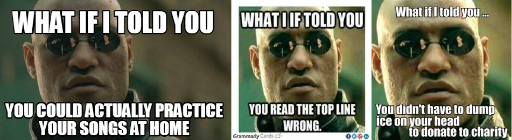}
    \end{minipage}

    \vspace{0.5cm} 
    
    \begin{minipage}{0.5\textwidth}
        \centering
        \includegraphics[width=\textwidth]{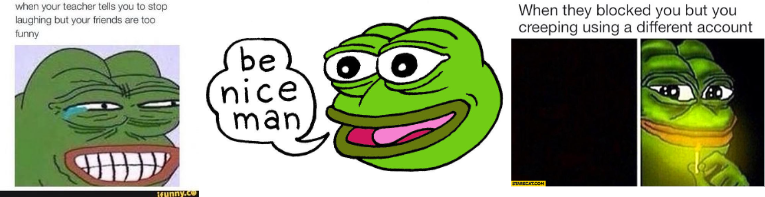}
    \end{minipage}

    \vspace{0.5cm} 
    
    \begin{minipage}{0.5\textwidth}
        \centering
        \includegraphics[width=\textwidth]{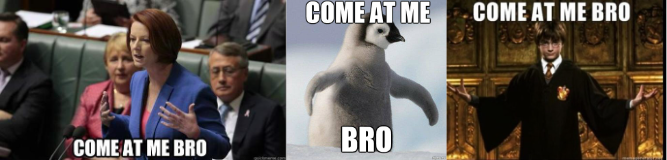}
    \end{minipage}

    \caption{Example of obtained results - groups of memes}
    \label{fig:memes}
\end{figure}

\begin{figure*}[htb]
    \centering
    \begin{subfigure}[b]{0.32\textwidth}
        \centering
        \includegraphics[width=\textwidth]{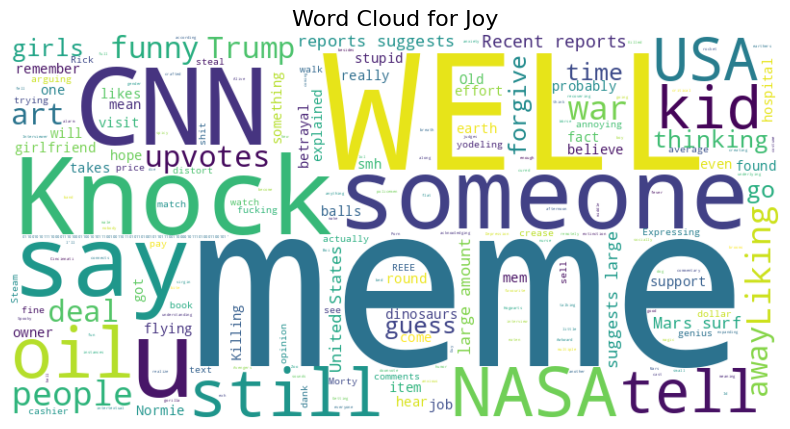}
        \caption{Joy}
        \label{fig:joy}
    \end{subfigure}
    \hfill
    \begin{subfigure}[b]{0.32\textwidth}
        \centering
        \includegraphics[width=\textwidth]{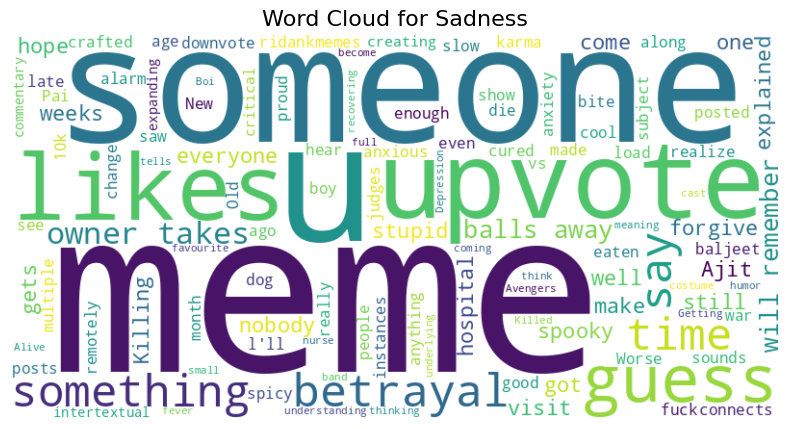}
        \caption{Sadness}
        \label{fig:sadness}
    \end{subfigure}
    \hfill
    \begin{subfigure}[b]{0.32\textwidth}
        \centering
        \includegraphics[width=\textwidth]{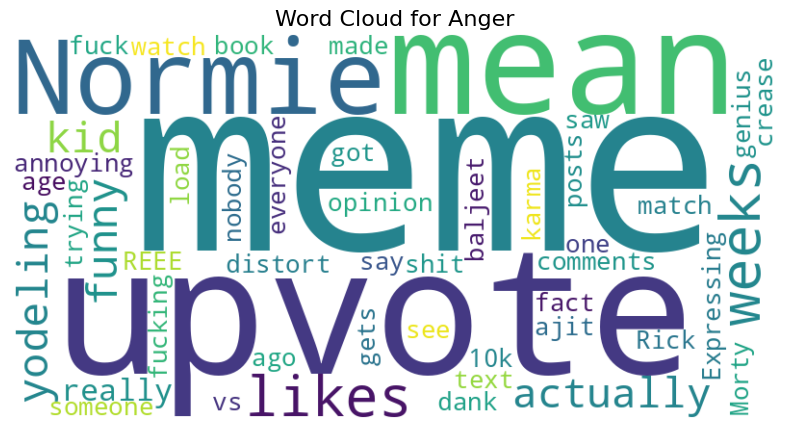}
        \caption{Anger}
        \label{fig:anger}
    \end{subfigure}

    \vfill
    \begin{subfigure}[b]{0.32\textwidth}
        \centering
        \includegraphics[width=\textwidth]{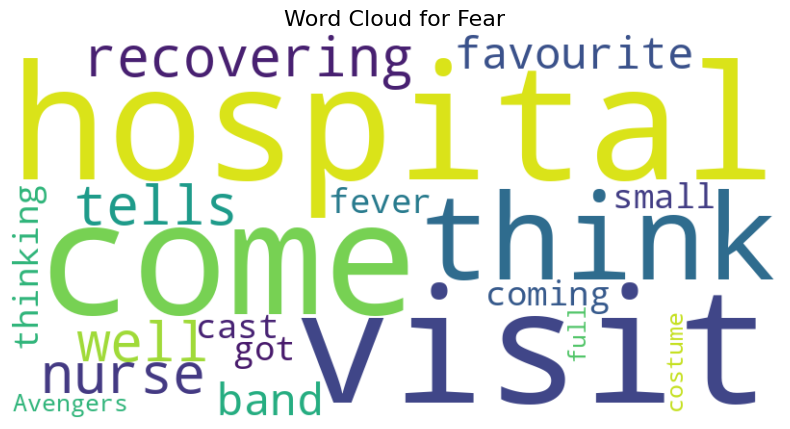}
        \caption{Fear}
        \label{fig:fear}
    \end{subfigure}
    \hfill
        \begin{subfigure}[b]{0.32\textwidth}
        \centering
        \includegraphics[width=\textwidth]{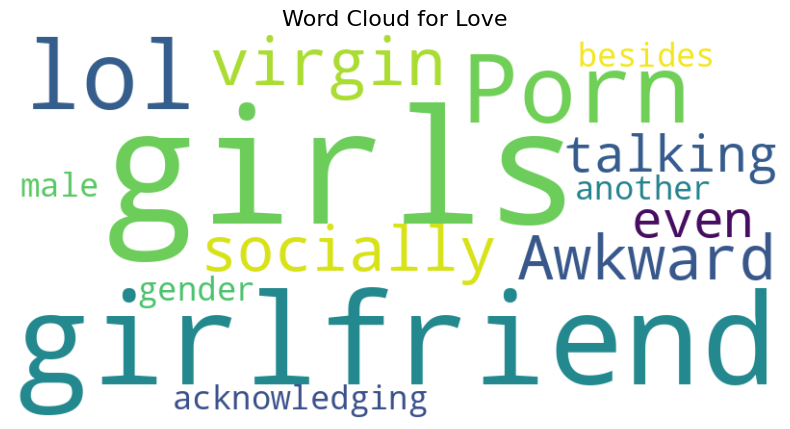}
        \caption{Love}
        \label{fig:love}
    \end{subfigure}
    \hfill
    \begin{subfigure}[b]{0.32\textwidth}
        \centering
        \includegraphics[width=\textwidth]{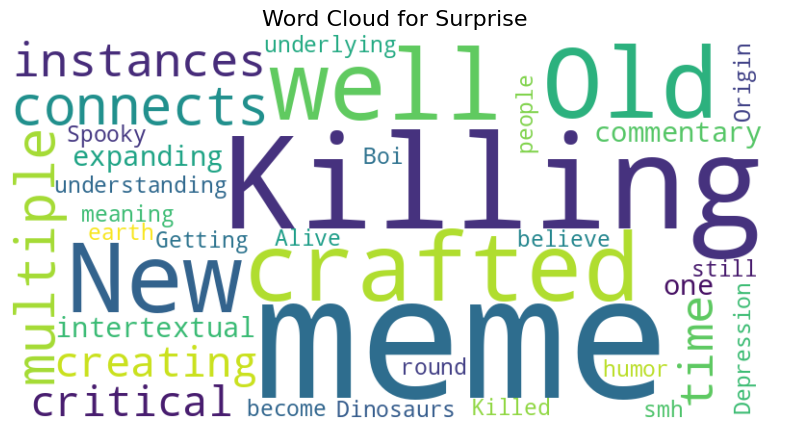}
        \caption{Surprise}
        \label{fig:surprise}
    \end{subfigure}
    \hfill
    \caption{Word clouds for different emotions detected in similar meme groups from Reddit Meme Dataset.}
    \label{fig:wordCloudEmotions}
\end{figure*}

\begin{equation} \text{Attention}(Q, K, V) = \text{softmax}\left(\frac{QK^T}{\sqrt{d_k}}\right) V \label{eq
} \end{equation}
where $Q$ is query, $K$ is key, and $V$ represents value matrices \cite{igali2024trackingemotionaldynamicschat}. In addition, $d_k$ refers to the dimensionality of the key vectors. The model can focus on information from different representation subspaces at distinct positions through multi-head attention \cite{igali2024trackingemotionaldynamicschat}:
\begin{equation} \text{MultiHead}(Q, K, V) = \text{Concat}(\text{head}_1, \dots, \text{head}_h)W^O \label{eq
} \end{equation}
\begin{equation} \text{head}_i = \text{Attention}(QW_i^Q, KW_i^K, VW_i^V) \label{eq
} \end{equation}
with $W_i^Q$, $W_i^K$, $W_i^V$, and $W^O$ being parameter matrices \cite{igali2024trackingemotionaldynamicschat}. Each transformer block in DistilBERT also contains a feed-forward network that applies transformations:
\begin{equation} \text{FFN}(x) = \text{max}(0, xW_1 + b_1)W_2 + b_2 \label{eq
} \end{equation}
To enhance training stability, DistilBERT employs layer normalization and residual connections:
\begin{equation} \text{LayerNorm}(x + \text{Sublayer}(x)) \label{eq
} \end{equation}

DistilBERT was trained on a 20 000 rows dataset derived from a collection of English Twitter messages to identify six basic emotions: \textit{joy, fear, anger, love, sadness} and \textit{surprise} \cite{igali2024trackingemotionaldynamicschat}, \cite{saravia-etal-2018-carer}. The model DistilBERT achieved an accuracy of 0.93 and was used to classify meme texts into six main emotions.

\section{Results}

\subsection{Example of grouping memes}

\begin{table*}[h]
\centering
\caption{Extracted image and text embeddings and cosine similarity scores for the Memotion dataset.  Higher similarity values indicate a closer semantic relationship between image-text pairs.
}
\resizebox{\textwidth}{!}{%
\begin{tabular}{c|c|c|c|c|c|c}
\toprule
\textbf{image\_name} & \textbf{text\_corrected} & \textbf{img\_emb} & \textbf{text\_emb} & \textbf{image\_1} & \textbf{...} & \textbf{image\_6992} \\
\midrule
\textbf{image\_1.jpg} & LOOK ... & [-5.74e-02 ... ] & [-5.74e-02 ... ] & tensor([[0.3461]]...) & ... & ... \\
\textbf{...} & ... & ... & ... & ... & ... & ... \\
\textbf{image\_6992.png} & The ... & [-1.07e-02 ...] & [1.24e-02 ...] & ... & ... & tensor([[0.8138]]...) \\
\bottomrule
\end{tabular}%
}
\label{table1}
\end{table*}

Fig. \ref{fig:memes} illustrates some of the grouping examples. From this, we can conclude that CLIP is not only a breakthrough in classification through unsupervised learning but also sets new standards for stability. This is because CLIP learns from image-text pairs that provide a richer context rather than just raw image classes. This highlights the importance of context not only in NLP but also in computer vision. 
Table~\ref{table1} presents an overview of the embeddings extracted for both image and text modalities in the Memotion dataset, along with their respective cosine similarity scores.  Higher similarity values indicate a closer semantic relationship between image-text pairs. 

\subsection{Performance Evaluation}
We conducted a user study with 51 participants aged 18-35 to evaluate our CLIP-based meme clustering model. Each participant viewed multiple meme clusters and indicated whether the memes seemed similar. The Arbitration and Agreement Rate (A:AR) metric was used to ensure consensus. The \textit{Agreement Rate} metric is calculated as the percentage of participants agreeing with the model’s groupings for each cluster. 

\begin{equation}
\text{Agreement Rate} = \frac{\text{Users Agree}}{\text{Total Users}} \times 100
\end{equation}

The overall Agreement Rate is then obtained by averaging the Agreement Rates across all clusters:

\begin{equation}
\text{Average Agreement Rate} = \frac{\sum \text{Agreement Rate}}{n}
\end{equation}

denoted by \( n \) — the number of meme clusters considered in the survey.

The Agreement Rate determination for each of the clusters varied between 45.01\% and 82.35\%, with an average of 67.23\%. This means that, on average, participants concurred with the model at a moderate level of 67.23\% in the manner in which the model grouped similar memes. The findings show that some clusters received more agreement, and this insight implies a good correlation between the model and the actual human perception corresponding to the clusters.

\begin{table*}[h]
\centering
\caption{Agreement Rates for Meme Groups in User Survey}
\resizebox{\textwidth}{!}{%
\begin{tabular}{c|c|c|c|c|c|c|c|c|c|c|c|c|c|c|c|c|c|c|c|c|c}
\toprule
\textbf{Meme group} & \textbf{1} & \textbf{2} & \textbf{3} & \textbf{4} & \textbf{5} & \textbf{6} & \textbf{7} & \textbf{8} & \textbf{9} & \textbf{10} & \textbf{11} & \textbf{12} & \textbf{13} & \textbf{14} & \textbf{15} & \textbf{16} & \textbf{17} & \textbf{18} & \textbf{19} & \textbf{20} & \textbf{21} \\
\midrule
\textbf{AR (\%)} & 64.71 & 45.10 & 70.59 & 52.94 & 60.78 & 70.59 & 66.67 & 82.35 & 78.43 & 60.78 & 80.39 & 70.59 & 56.86 & 58.82 & 66.67 & 68.63 & 76.47 & 74.51 & 60.78 & 74.51 & 70.59 \\
\bottomrule
\end{tabular}%
}
\label{tab:agreement_rates}
\end{table*}

\color{Black}

\subsection{Emotion Analysis Results}
Each meme was analyzed for its textual content using DistilBert, and the detected emotions were visualized using a cloud of words, where each cloud represents the most common words associated with a particular emotion (see Fig.~\ref{fig:wordCloudEmotions}). For example:
\begin{itemize}
    \item \textbf{Joy}. The word cloud highlights positive and uplifting terms, reflecting happiness in memes (Fig.~\ref{fig:wordCloudEmotions}a).
    \item \textbf{Sadness}: The cloud is dominated by words associated with a sad or melancholic context (Fig.~\ref{fig:wordCloudEmotions}b).
    \item \textbf{Anger}: Terms expressing frustration or conflict are highlighted (Fig.~\ref{fig:wordCloudEmotions}c).
\end{itemize}

\begin{table}[tb]
\centering
\caption{Emotion Distribution in the Dataset}
\scriptsize  
\begin{tabular}{l|r|r|r|r|r|r}
\toprule
\textbf{Emotion} & \textbf{Joy} & \textbf{Anger} & \textbf{Fear} & \textbf{Sadness} & \textbf{Surprise} & \textbf{Love} \\
\midrule
\textbf{Count}       & 3120  & 2374  & 629   & 627   & 143   & 94    \\
\textbf{Percent (\%)} & 44.65 & 33.98 & 9.00  & 8.97  & 2.05  & 1.35  \\
\bottomrule
\end{tabular}%
\label{tab:emotion_distribution}
\end{table}

Table \ref{tab:emotion_distribution} illustrates the distribution of emotions within the Memotion dataset, revealing that joy (44.65\%) is the most prevalent emotion, followed by anger (33.98\%), while love (1.35\%) and surprise (2.05\%) are the least represented. These findings highlight the dominance of positive and intense emotions in meme content.

In addition, we performed analysis to identify the relationship between emotions and meme content (motivational, sarcastic, offensive, humor, sentiment). As a result, we identified a statistically significant relationship between emotion and motivational content (p-value 0.045). This suggests that motivational content tends to evoke stronger emotional responses.

\section{Limitations and Future Work}


While our study demonstrates promising results, several limitations remain, like limited user study scope, multimodal ambiguities, evolving nature of memes, and dataset biases.

Due to rapidly changing trends, memes can quickly become outdated. A model trained on static datasets may not perform well on newly emerging memes. In our analysis, we found that the model occasionally performed poorly due to memes' subtle and often ambiguous meanings. The Memotion and Reddit Meme datasets used in this study follow specific meme formats in the English language and may not fully represent the diversity of meme styles found on various platforms.

The findings revealed that the agreement rates ranged between 45.10\% and 82.35\% through clusters. This implies that the model performs poorly in handling some types of memes, probably because of variations in the positioning of texts and images. Additionally, some memes in the dataset lack text, which our emotion analysis requires. 

Only 51 participants aged 18-35 were included in the survey, so it is uncertain how various user demographics perceive memes. Variations in meme styles, languages, or cultural aspects that are not captured in the presented dataset may decrease the generalization of the findings.

Further research can be dedicated to improving memes' text and image features. 
It can explore advanced cross-modal fusion techniques, such as attention-based fusion mechanisms, to better integrate textual and visual features. We also plan to incorporate external knowledge sources like social media trends. In addition, more diverse participants’ ages and cultural backgrounds must be involved. This diversity will improve the assessment of model performance across the different segments, helping us better represent the complex nature of memes.




\section{Conclusion}
 In this research, we aimed to develop a model for multi-modal meme analysis. Our study results show that the model effectively groups similar memes based on their visual and textual content.

 Memes are often perceived as humorous, yet they contain a broad spectrum of emotions. For instance, motivational memes may draw from emotions such as anger or happiness. Additionally, we aimed to make a proposal for a list of basic emotions that memes produce or provoke. Results show that emotions of \textit{anger} and \textit{joy} are dominant in memes. Our findings also show that motivational memes tend to evoke stronger emotional responses. Such a perspective contributes to further research on the emotional impact of internet memes.


 A number of recent studies highlighted the effectiveness of multimodal pre-training and cross-modal interaction modeling in detecting and understanding hateful memes. Some of them have focused on the analysis of memes using CLIP and consideration of both visual and textual information. \cite{Gokul2022Hate} proposed the Hate-CLIPper architecture, which uses CLIP encoders and a feature interaction matrix to model cross-modal interactions. Similarly, \cite{Y2022On} used CLIP for multimodal contrastive learning to examine how hateful memes developed. 
 
 

Potential applications include Content Moderation in social media (e.g., Detecting Offensive Memes), Meme Search and Retrieval, Personalized Meme Recommendations, and Meme Trend Detection, among others.

\section*{Acknowledgement}
This research has been funded by the Science Committee of the Ministry of Science and Higher Education of the Republic of Kazakhstan (Grant No. AP22786412)

\bibliography{library.bib}
\end{document}